\documentclass[10pt,twocolumn,letterpaper,capitalize]{article}

\usepackage{cvpr}             
\usepackage{comment}

\usepackage[utf8]{inputenc} 
\usepackage[T1]{fontenc}    
\usepackage{url}            
\usepackage{nicefrac}       
\usepackage{microtype}      
\PassOptionsToPackage{usenames,dvipsnames}{xcolor}
\usepackage{amsfonts}       
\usepackage{lmodern}
\usepackage{amssymb}        

\usepackage{booktabs}   
\usepackage{tabularx}   
\usepackage{colortbl}   
\usepackage{multirow}
\usepackage{makecell}

\newcommand{\mc}[3]{\multicolumn{#1}{#2}{#3}}

\usepackage{wrapfig}
\usepackage{mwe} 

\usepackage[font=normal,labelfont=bf]{caption}
\usepackage[font=normal]{subcaption}

\usepackage[normalem]{ulem} 
\usepackage{array}

\usepackage{xparse}
\usepackage{pifont}

\usepackage{ifthen}

\definecolor{cvprblue}{rgb}{0.21,0.49,0.74}
\usepackage[pagebackref,breaklinks,colorlinks,citecolor=cvprblue,linkcolor=red!75!black]{hyperref}

\usepackage[capitalize]{cleveref}
\crefformat{section}{Section~#2#1#3}
\crefformat{subsection}{Section~#2#1#3}
\crefformat{subsubsection}{Section~#2#1#3}
\crefformat{figure}{Fig.~#2#1#3}
\crefformat{equation}{Eq.~#2#1#3}
\crefformat{table}{Table~#2#1#3}
\crefformat{algorithm}{Alg.~#2#1#3}
\usepackage{algpseudocode}
\usepackage[linesnumbered,ruled,vlined]{algorithm2e}

\SetCommentSty{mycommfont}
\SetKwInput{KwInput}{Input}

\usepackage{enumitem}

\usepackage{xfp}

\newlength\savewidth\newcommand\shline{\noalign{\global\savewidth\arrayrulewidth
  \global\arrayrulewidth 1pt}\hline\noalign{\global\arrayrulewidth\savewidth}}
\newlength\thinwidth

\definecolor{Gray}{gray}{0.93}
\newcolumntype{a}{>{\columncolor{Gray}}c}
\newcommand{\colorrow}{\rowcolor{blue!5}}

\usepackage{xspace} 
\makeatletter
\DeclareRobustCommand\onedot{\futurelet\@let@token\@onedot}
\def\@onedot{\ifx\@let@token.\else.\null\fi\xspace}
 
\def\ie{\emph{i.e}\onedot}

\makeatother

\usepackage{amsmath,amsfonts,bm}

\def\1{\bm{1}}

\def\vmu{{\bm{\mu}}}

\def\vx{{\bm{x}}}

\def\vz{{\bm{z}}}

\def\mX{{\bm{X}}}
\def\mY{{\bm{Y}}}
\def\mZ{{\bm{Z}}}

\DeclareMathAlphabet{\mathsfit}{\encodingdefault}{\sfdefault}{m}{sl}
\SetMathAlphabet{\mathsfit}{bold}{\encodingdefault}{\sfdefault}{bx}{n}

\def\gB{{\mathcal{B}}}

\def\gD{{\mathcal{D}}}
\def\gE{{\mathcal{E}}}

\def\gL{{\mathcal{L}}}
\def\gM{{\mathcal{M}}}

\def\gV{{\mathcal{V}}}

\def\gX{{\mathcal{X}}}

\newcommand{\E}{\mathbb{E}}

\def\paperID{3847} 
\def\confName{CVPR}
\def\confYear{2025}

\title{From Prototypes to General Distributions: An Efficient Curriculum for Masked Image Modeling}

\author{Jinhong Lin\thanks{Equal Contribution}~~~Cheng-En Wu\footnotemark[1]~~~Huanran Li~~~Jifan Zhang~~~Yu Hen Hu~~~Pedro Morgado  \\
University of Wisconsin–Madison \\
{\tt\small \{jlin398, cwu356, hli488, jzhang769, yhhu, pmorgado\}@wisc.edu}}

\begin{document}

\maketitle

\newcommand{\TemperatureExp}{
\begin{tabular}{cccc}
    \toprule
    $\tau$ & $|\gD_\tau| / |\gD|$ & \bf NN & \bf LP  \\
    \midrule
    0.01 & 0.004 & 5.69 & 18.22  \\
    0.05 & 0.122 & 29.51 & 46.38  \\
    0.08 & 0.236 & 39.67 & 61.86  \\
    0.2 & 0.495 & 41.57 & 67.71  \\
    0.4 & 0.592 & 36.86 & 67.15  \\
    0.6 & 0.614 & 35.17 & 66.45  \\
    inf & 0.632 & 31.70 & 64.94  \\
    \midrule
    \mc{2}{c}{Annealed} & 47.40 & 68.84  \\
    \bottomrule
\end{tabular}
}

\newcommand{\AlgorithmExp}{
\begin{tabular}{lccc}
    \toprule
    \textbf{Representation Space} & \bf NN & \bf LP \\
    \midrule
    SIFT~\cite{davies1979cluster} & 36.85 & 69.17 \\
    VIT-B/16 Untrained & 32.63 & 65.65 \\
    VIT-B/16 MAE~\cite{he2021masked} & 34.14 & 66.60 \\
    VIT-B/16 DiNO~\cite{caron2021emerging} & 40.15 & 68.73 \\
    VIT-B/16 Supervised~\cite{dosovitskiy2020image} & 39.70 & 68.09 \\
    \bottomrule
\end{tabular}
}
\newcommand{\ArchitectureExp}{
\begin{tabular}{lccc}
    \toprule
    \textbf{Architecture} & \bf NN & \bf LP \\
    \midrule
    ViT B/16 & 40.15 & 68.73 \\
    ViT S/16 & 41.10 & 69.00 \\
    ViT S/8 & 41.05 & 69.23 \\
    \bottomrule
\end{tabular}
}

\newcommand{\NumClusterExp}{
\begin{tabular}{ccc}
    \toprule
    \bf \# Clusters & \bf NN & \bf LP \\
    \midrule
    100  & 38.64 & 68.54 \\
    500 & 40.76 & 68.51  \\
    1000 & 41.05 & 69.23  \\
    1500 & 39.65 & 68.35 \\
    2000 & 38.90 & 68.27 \\
    \midrule
    978 & 41.05 & 69.00 \\
    \midrule
    Oracle (1000) & 37.83 & 68.49 \\
    \bottomrule
\end{tabular}
}

\newcommand{\MainExp}{
\begin{tabular}{lcccc}
    \toprule
    \textbf{Curriculum} & \bf Epochs & \bf NN & \bf LP & \bf FT \\
    \midrule
    None (MAE)~\cite{he2022masked} & 800  & 28.73 & 64.25 & 83.08 \\
    Hard-samples~\cite{sorscher2022beyond} & 800 & 24.63 & 62.09 & 82.95 \\
    \midrule
    \colorrow{}%
    Prototypical (Ours) & 800 & 47.40 & 68.84 & 83.31 \\
    \bottomrule
\end{tabular}
}

\newcommand{\VarEpoch}{
\begin{tabular}{cccccc}
    \toprule
    \textbf{Curriculum} & \textbf{Epoch} & \bf NN & \bf LP & \bf FT \\
    \midrule
    \multirow{4}{*}{\centering None (MAE~\cite{he2022masked})} & 200  & 24.88 & 59.31 & 82.46 \\
                                       & 400  & 28.73 & 62.08 & 82.91 \\
                                       & 800  & 30.25 & 64.25 & 83.08 \\
    \midrule
    \multirow{3}{*}{\centering Prototypical (Ours)} & 200  & 34.92 & 63.74 & 82.75 \\
                                       & 400  & 38.09 & 66.34 & 83.09 \\
                                       & 800  & \bf 47.40 & \bf 68.84 & \bf 83.31 \\
    \bottomrule
\end{tabular}
}

\begin{abstract}
    Masked Image Modeling (MIM) has emerged as a powerful self-supervised learning paradigm for visual representation learning, enabling models to acquire rich visual representations by predicting masked portions of images from their visible regions. While this approach has shown promising results, we hypothesize that its effectiveness may be limited by optimization challenges during early training stages, where models are expected to learn complex image distributions from partial observations before developing basic visual processing capabilities. To address this limitation, we propose a prototype-driven curriculum learning framework that structures the learning process to progress from prototypical examples to more complex variations in the dataset. Our approach introduces a temperature-based annealing scheme that gradually expands the training distribution, enabling more stable and efficient learning trajectories. Through extensive experiments on ImageNet-1K, we demonstrate that our curriculum learning strategy significantly improves both training efficiency and representation quality while requiring substantially fewer training epochs compared to standard Masked Auto-Encoding. Our findings suggest that carefully controlling the order of training examples plays a crucial role in self-supervised visual learning, providing a practical solution to the early-stage optimization challenges in MIM.
    \vspace{-3ex}
\end{abstract}

\begin{figure}[t]
\centering
\includegraphics[width=1.0\linewidth]{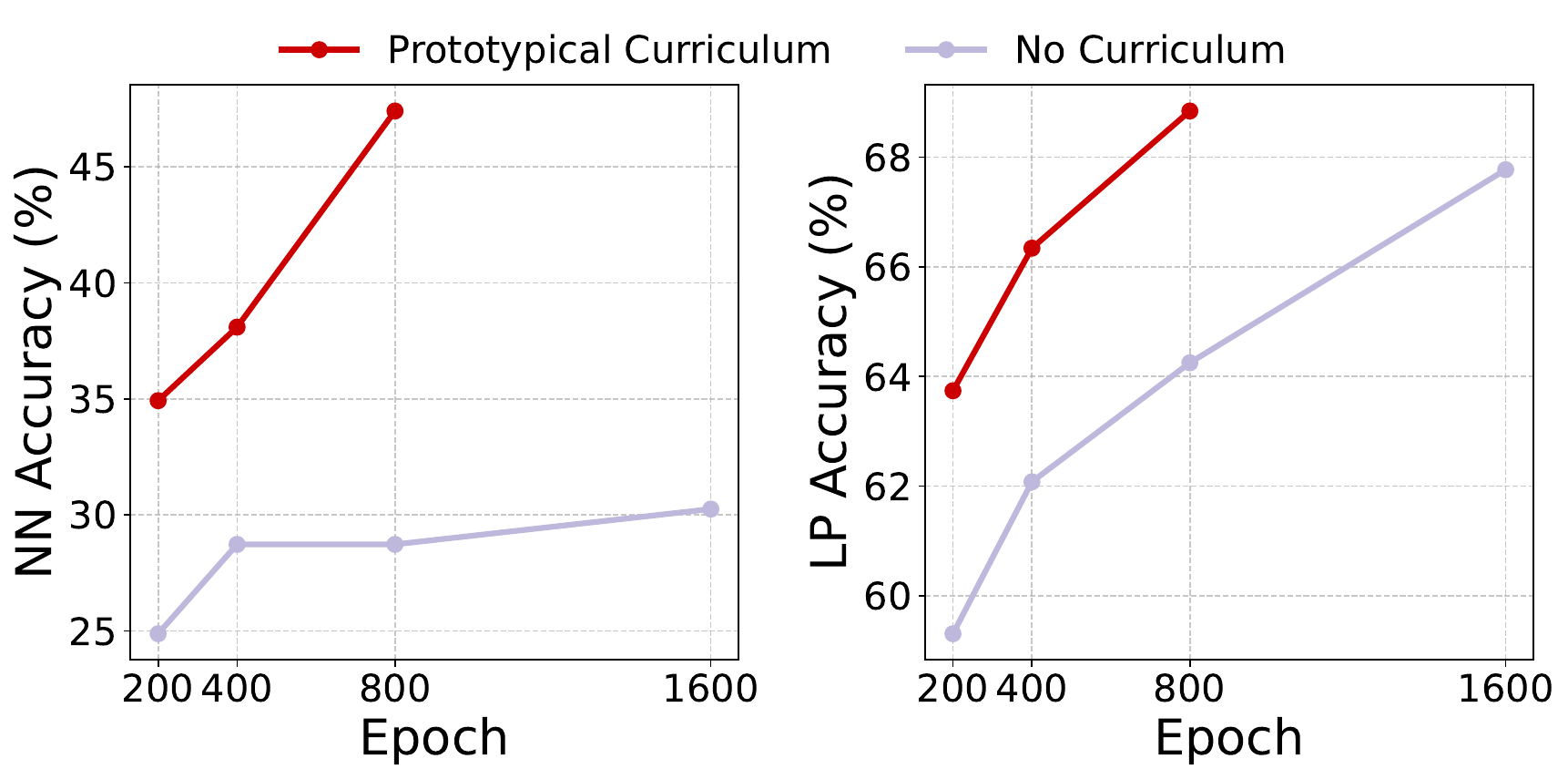}
\vspace{-4ex}
\caption{Our prototypical curriculum (red line) significantly improves MAE downstream performance on ImageNet-1k compared to the standard MAE training strategy (purple line). We show both the nearest neighbor (left) and linear probe (right) accuracy of models trained for different numbers of epochs.}%
\label{fig:teaser}
\vspace{-3ex}
\end{figure}

\section{Introduction}%
\label{sec:intro}
Masked Image Modeling (MIM), initially explored in~\cite{vincent2010stacked, pathak2016context} and popularized in~\cite{he2021masked}, has emerged as a powerful self-supervised learning paradigm for visual representation learning. By learning to predict masked portions of images from their visible regions, these models can acquire rich visual representations without requiring explicit labels. However, despite their success, current MIM approaches face a fundamental challenge that has received limited attention: the optimization difficulties during early training stages.

The core challenge lies in the immediate expectation for models to learn a complex mapping from partial observations to the full distribution of natural images, even before developing basic visual processing capabilities. This is analogous to expecting an art student to reproduce complex masterpieces before mastering fundamental sketching techniques. This difficulty is compounded by the inherent complexity and high dimensionality of natural image distributions, making the initial learning phase particularly challenging and potentially inefficient.

In this paper, we propose a novel curriculum learning framework for MIM that addresses these optimization challenges while maintaining the core reconstruction objective. Our approach is motivated by the observation that human visual learning often progresses from simple to complex patterns \cite{sheybani2024curriculum,orhan2020self}, and that certain prototypical examples can serve as effective ``templates'' for learning basic visual concepts. Rather than immediately exposing the model to the full complexity of natural image distributions, we structure the learning process to begin with a simplified distribution of prototypical examples that capture essential visual patterns for each semantic category. Our key insight is that by carefully controlling the expansion of the training distribution from prototypical examples to more complex variations, we can create a more efficient and stable learning trajectory, ultimately leading to more robust representation learning.

Through extensive experiments on the ImageNet-1K dataset, we demonstrate that incorporating a prototype-driven curriculum strategy significantly improves masked image modeling, both in terms of training efficiency and representation quality (\cref{fig:teaser}). By gradually transitioning from prototypical to diverse examples during training, our approach achieves superior performance while requiring substantially fewer training epochs compared to standard Masked Auto-Encoding (MAE)~\cite{he2022masked}. Extensive ablation studies also reveal that the success of our method stems from careful prototype identification and a progressive learning schedule, suggesting that the order in which examples are presented plays a crucial role in self-supervised visual learning.

\section{Related Work}
\label{sec:related-work}

\paragraph{Masked Image Modeling}
Masked Image Modeling has emerged from early self-supervised learning approaches in computer vision, gaining prominence after the success of BERT in natural
language processing. While initial attempts at masking in vision were made in~\cite{vincent2010stacked,pathak2016context} using convolutional architectures for
denoising and inpainting tasks, modern MIM approaches took shape with the advent of Vision Transformers. Two influential concurrent works,
MAE~\cite{he2022masked} and SimMIM~\cite{xie2022simmim}, established the core framework by introducing high-ratio random masking and transformer-based
architectures. This foundation sparked numerous innovations: some methods focused on improving the masking strategy through
attention-guided~\cite{kakogeorgiou2022hide} or semantically-driven approaches~\cite{li2022semmae}, while others explored alternative learning objectives by
combining reconstruction with contrastive learning~\cite{chen2022sdae} or incorporating knowledge distillation~\cite{huang2023contrastive}. MIM has since
evolved beyond static images to handle various modalities including video~\cite{tong2022videomae, wang2023videomae}, point clouds~\cite{zhang2022point,
pang2022masked}, and multi-modal data~\cite{mo2024unveiling, huang2024mavil, bachmann2022multimae}, demonstrating its versatility as a self-supervised learning
paradigm. Our work builds on this foundation by proposing a novel data selection strategy that accelerates MIM training by emphasizing common data early in the
learning process.

\paragraph{Dataset Selection}
Dataset selection refers to the process of selecting a subset of the data that best represents the entire dataset, enabling efficient training without
sacrificing performance. Data selection techniques in supervised learning focus on identifying key subsets of training data that have the greatest impact on
model performance. One research avenue, most related to our method, explores proxy models to locate informative subsets. For instance, studies such
as~\cite{birodkar2019semantic, ghorbani2019data, coleman2019selection, sorscher2022beyond} employ k-means clustering or uncertainty estimates from proxy models
to select data based on feature space distance.  Other methods leverage sample difficulty and uncertainty.~\cite{toneva2018empirical} found that frequently
forgotten examples during training are more valuable for model refinement. Gradient-based selection techniques have also been used in several studies,
including~\cite{mirzasoleiman2020coresets,pooladzandi2022adaptive,killamsetty2021grad,katharopoulos2018not,paul2021deep}, to identify data points that closely
approximate the overall gradient, facilitating efficient subset selection based on gradient insights. In general, these methods show that to retain performance
in supervised learning, the selected subset must be representative of the visual diversity in the dataset, often undersampling easy to learn ``prototypical''
samples, which we show are beneficial in the early phases of MIM training to accelerate learning. 

In self-supervised learning (SSL), dataset selection has been less explored, though recent advances are addressing this gap. Geometry-based methods, such as
those proposed by~\cite{joshi2023data}, leverage augmentation similarities among data points within a dataset to enhance data efficiency in contrastive
learning. However, this method suffers from scalability issues and relies on a strongly pre-trained model to extract fine-grained features. In contrast,
our proposed approach eliminates this dependency, allowing for data selection based solely on low-level features.
Dynamic selection methods, such as~\cite{tripathidynamic}, adapt data selection dynamically during training by refining the selection process from coarse to fine-grained as the model evolves. While effective for contrastive learning, this approach necessitates frequent and substantial computational overhead. Our method, on the other hand, is tailored specifically for Masked Image Modeling (MIM) and avoids additional computational demands during training.%

\vspace{-2ex}
\paragraph{Importance Sampling Techniques}
Importance sampling has emerged as a promising technique for data selection in stochastic optimization. Early methods by \cite{alain2015variance,
katharopoulos2018not, loshchilov2015online, schaul2015prioritized} prioritized samples with higher importance, often determined by large gradient norms or high
loss values. These techniques aim to reduce variance in stochastic gradients, occasionally improving convergence rates. However, they generally lack rigorous
convergence guarantees or substantial speedups when applied to overparameterized models, such as deep neural networks. More recent work
by~\cite{mindermann2022prioritized} refines these concepts by selecting non-noisy, non-redundant, and task-relevant examples that minimize loss on a holdout
validation set. While this approach accelerates training, it depends on a validation set and lacks theoretical convergence guarantees for non-convex models,
limiting its utility in deep learning. 

In gradient approximation, the CRAIG method~\cite{mirzasoleiman2020coresets} selects sample subsets that approximate the total gradient sum by maximizing a
submodular function. While CRAIG guarantees convergence for strongly convex models, its effectiveness declines in non-convex settings like deep learning.
GradMatch~\cite{killamsetty2021grad} and Glister~\cite{killamsetty2021grad} further explore gradient-based subset selection. GradMatch uses orthogonal matching
pursuit (OMP) to select minimal training subsets, while Glister employs a validation set for gradient alignment. Both methods are effective in smaller scenarios
but face scalability challenges with large datasets.%
\vspace{-4pt}
\begin{figure*}[t]
    \centering
    \includegraphics[width = 1\textwidth]{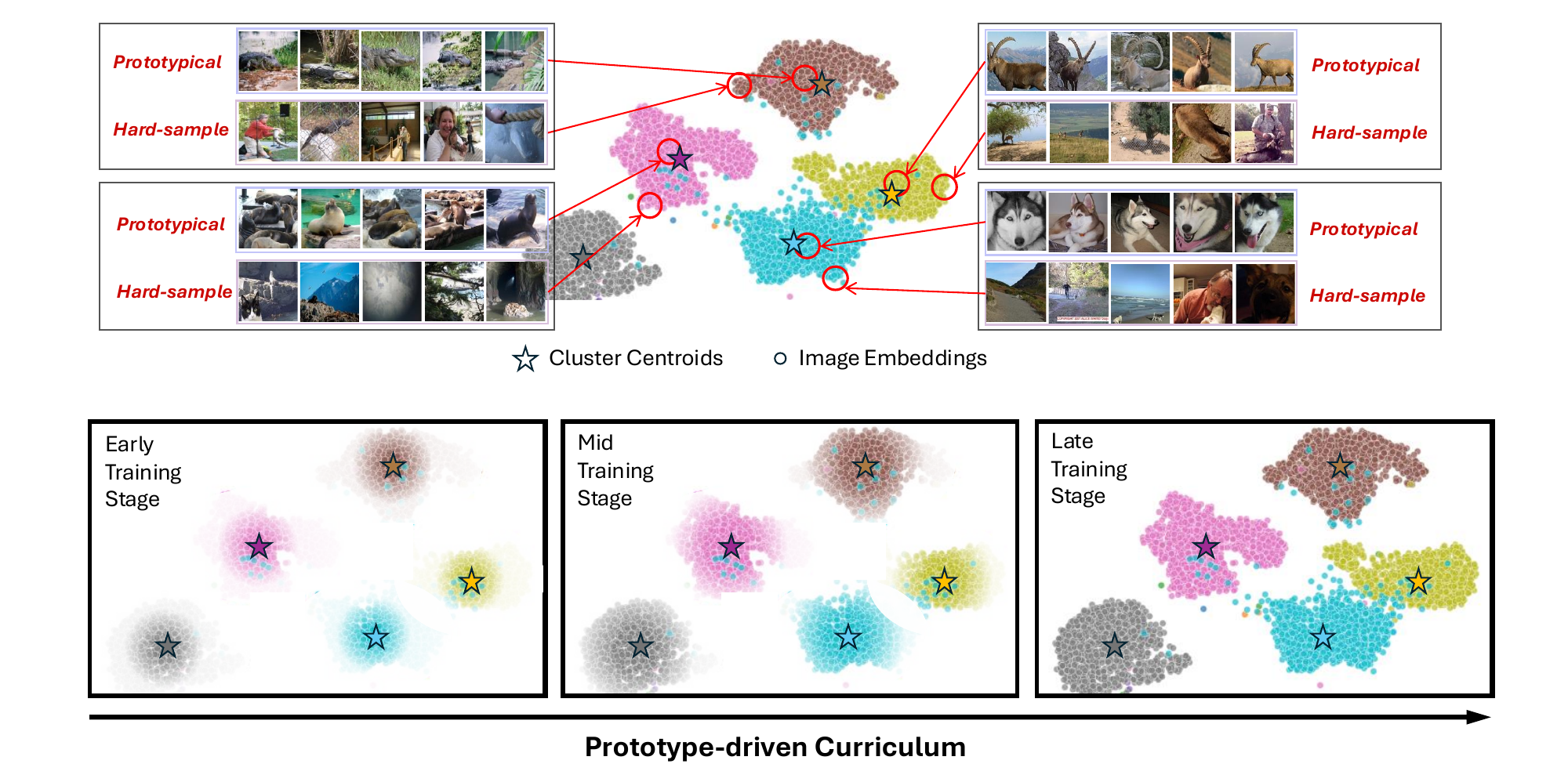}
    \vspace{-6ex}
    \caption{{\bf Framework Overview.} We introduce a curriculum learning framework for Masked Autoencoders (MAE) that prioritizes sampling specific images early in training and progressively expands to cover the entire dataset. K-means clustering is applied within a visual feature space to identify clusters that exhibit prototypical visible semantics. The framework selects images near the cluster centers as prototypes, representing basic and common visual features, while more challenging examples are sampled from the cluster boundaries. Additional visualizations are provided in~Fig.\ref{fig:viz_cluster_samples}.}%
    \label{fig:framework}
    \vspace{-4ex}
\end{figure*}

\section{Method}
We propose a curriculum learning strategy to enhance training of Masked Image Modeling methods, in particular, Masked Auto Encoders (MAE)~\cite{he2022masked}. An overview of our strategy is illustrated in Figure~\ref{fig:framework}.

\vspace{-4pt}
\subsection{[Background] Masked Image Modeling}
Masked Image Modeling (MIM) is a self-supervised learning framework that seeks to reconstruct images from partial observations. 
Given an input image $x \in \Re^{H \times W \times 3}$, MIM procedures first divide it into $N$ non-overlapping patches, $\mX = [\vx_1, \vx_2, \ldots, \vx_N]$, each of size $\vx_i \in \Re^{P \times P \times 3}$. Then, given a pre-specified mask ratio $m$ (typically 75\%), $\mX$ is randomly partitioned into two disjoint sets of visible $\gV$ and masked patches $\gM$, where $|\gM| = Nm$ and $|\gV| = N(1-m)$. 
The visible patches are processed by an encoder model $\gE(\cdot)$ to encode them into visible latent representations $\mZ_\gV = \gE(\mX_\gV;\theta_\gE) \in \Re^{N(1-m) \times d}$, where $d$ denotes the feature dimensionality and $\theta_\gE$ the encoder parameters.
The visible representations $\mZ_\gV$ together with the positions of masked patches $p_\gM$ are then used to reconstruct the full sequence of patches by a decoder model, $\mY = \gD(\mZ_\gV, p_\gM; \theta_\gD)\in \Re^{N \times (P \times P \times 3)}$ parameterized by $\theta_\gD$. Both encoder and decoder are trained to minimize the mean squared error between the predicted and actual pixel values of masked patches
\begin{equation}
    \gL = \E_{\gX}\left[\left\|\mX_{\gM} - \mY_{\gM}\right\|_2^2\right].
\end{equation}
After pre-training, only the encoder $\gE$ is retained for downstream tasks, while the decoder $\gD$ is discarded. 

\subsection{Improving MIM through Prototypical Curriculum Learning}
Despite the success of MIM approaches in self-supervised visual representation learning, these models face significant optimization challenges during the early stages of training. This challenge stems from attempting to learn the complex mapping from partial observations to the full distribution of natural images, even before the model has developed basic visual processing capabilities. This is analogous to asking a student to reproduce complex artwork before mastering basic sketching techniques. To address this limitation, we propose a curriculum learning approach that, without changing the MIM reconstruction objective, aids the learning process by starting from a simplified distribution of training images, and gradually broadening the training distribution to include more complex examples.

Our key insight is that the training process can be made more efficient by initially focusing on prototypical examples that capture essential visual patterns for each semantic category (\cref{fig:framework}). These prototypical images serve as ``templates'' that are easier to reconstruct, allowing the model to first learn basic visual concepts before tackling more complex variations. As training progresses, we gradually expand the distribution to include examples that deviate further from these prototypes, eventually covering the full range of natural image variations. 
As we will show empirically, this structured learning trajectory leads to more stable optimization, and improved training efficiency through structured exploration of the data distribution. 

\subsubsection{Identifying Prototypical Examples}
Prototypical images can be identified through clustering in feature space. Specifically, we cluster image representations $\vz_i\in\Re^{d}$ using the mini-batch $K$-means clustering, and leverage the resulting $K$ cluster centroids $\vmu_k\in\Re^{d}$ to locate likely prototypical images, \ie, images whose representations $\vz_i$ are close to each centroid $\vmu_k$.
\vspace{-2ex}
\paragraph{Choosing the Number of Clusters}
\label{para:DB}
To automate the choice of $K$, we vary it in the range $2 \leq K \leq K_\textit{max}$, and use the Davies-Bouldin index $\gD\gB$ to assess the clustering quality. $\gD\gB$ measures both how compact each cluster is and how well-separated they are from other clusters~\cite{davies1979cluster}.
\vspace{-2ex}

\paragraph{Representation space}
The choice of representation space can significantly impact the selection of prototypical images. To study its impact on the curriculum, we investigate representations across different levels of semantic abstraction, from low-level visual descriptors such as traditional computer vision features (e.g., SIFT, HOG) to representations obtained from pretrained models like DINO~\cite{caron2021emerging} or MAE~\cite{he2022masked}. We find that while prototype-driven curricula provide benefits regardless of feature space, higher-level semantic features from self-supervised models like DINO are the most effective.

\subsubsection{Prototypical Curriculum}
The proposed curriculum controls the emphasis on prototypical examples through a temperature-based sampling mechanism. For each data point $\vz_i$, we first quantify its ``prototypicality'' based on the distance to the nearest centroid ($\vmu_k$).
\begin{equation}
    d(\vz_i, \vmu_k) = \|\vz_i - \vmu_k\|_2
\end{equation}
However, the divergence within each clusters may vary, with some clusters being tightly packed around their center (low divergence) and others more spread out (high divergence). To prevent sampling from being biased by variations in cluster divergences, we normalize the distances within each cluster using min-max normalization
\begin{equation}
    \hat{d}_i = \frac{d(\vz_i, \vmu_k)-d_\text{min}}{d_\text{max} - d_\text{min}},
\end{equation}
where $d_{\text{max}}$ and $d_{\text{min}}$ are the maximum and minimum distances within cluster $k$. Training data is then sampled from $\gD$ with replacement according to the following temperature-controlled softmax probabilities
\begin{equation}
    P(x_i,\tau) = \frac{\exp(-\hat{d}_i/\tau)}{\sum_j \exp(-\hat{d}_j/\tau)},
\end{equation}
where $\tau$ is the temperature parameter that controls the sharpness of the distribution. 
In the early stages of training, $\tau$ is low to concentrate the sampling distribution around the cluster centers and prioritize prototypical examples. As training progresses, $\tau$ is gradually increased, allowing the model to encounter more diverse examples while building on its foundational understanding of prototypical patterns.
\subsubsection{Annealing Temperature by Scheduling the Effective Data Size}
While temperature-controlled sampling offers a flexible mechanism to adjust the curriculum over time, defining appropriate temperature values ($\tau$) can be challenging due to their strong dependence on dataset characteristics. A more intuitive approach is to directly specify ``effective dataset sizes'', \ie, what portion of the dataset the model should be exposed to at each training stage.

\begin{figure}[h]
    \centering
    \includegraphics[width=\linewidth]{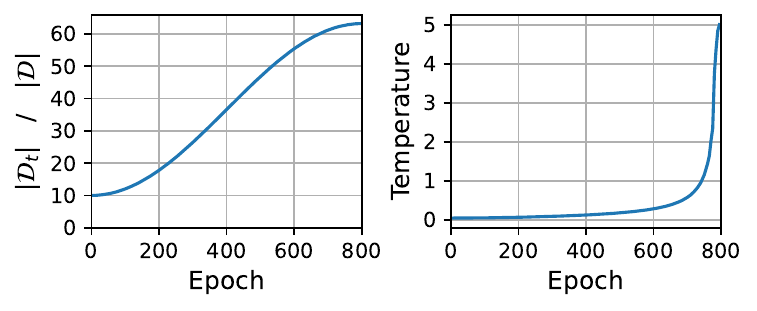}
    \vspace{-16pt}
    \caption{{\bf Temperature annealing schedule and its effect on training distribution.} The temperature parameter $\tau$ gradually increases over training epochs, controlling the breadth of the sampled distribution. The effective dataset size, measured as $|\gD_\tau| / |\gD|$, shows the corresponding fraction of the training data that is sampled in a single epoch at each temperature level. As the temperature increases the effective dataset size approaches the theoretical limit of $1-1/e$.}%
    \label{fig:temperature}
    \vspace{-4ex}
\end{figure}

We define the effective dataset size $|\gD_\tau|$ as the number of unique samples drawn from $\gD$ over one epoch when sampling with replacement according to the distribution $P(\cdot, \tau)$, which can be expressed as
\begin{equation}
    \label{eq:effective-size}
    |\gD_\tau| = \sum_{i=1}^{|\gD|} \left[1 - \left(1 - P\left(x_i, \tau\right)\right)^{|\gD|}\right].
\end{equation}
The curriculum is implemented by gradually increasing $|\gD_\tau|$ during training using a cosine schedule. This schedule spans from a pre-specified initial value to its theoretical limit
\begin{equation}
    |\gD| \times\lim_{\substack{\\\tau \to \infty\\|\gD|\to\infty}} \frac{|\gD_\tau|}{|\gD|} = |\gD|\times(1-1/e).
\end{equation}
For any desired $|\gD_\tau|$, we can determine the corresponding temperature $\tau$ by solving \cref{eq:effective-size}. Although this equation lacks a closed-form solution, we can efficiently solve for $\tau$ using binary search, taking advantage of the monotonic relationship between $|\gD_\tau|$ and $\tau$.~\cref{fig:temperature} illustrates this temperature annealing process.

\section{Experiments}%
\label{sec:experiments}

\subsection{Experimental Setup}%
\label{sec:study-design}

\begin{figure*}[t]
    \centering
    \includegraphics[width = 0.9\textwidth]{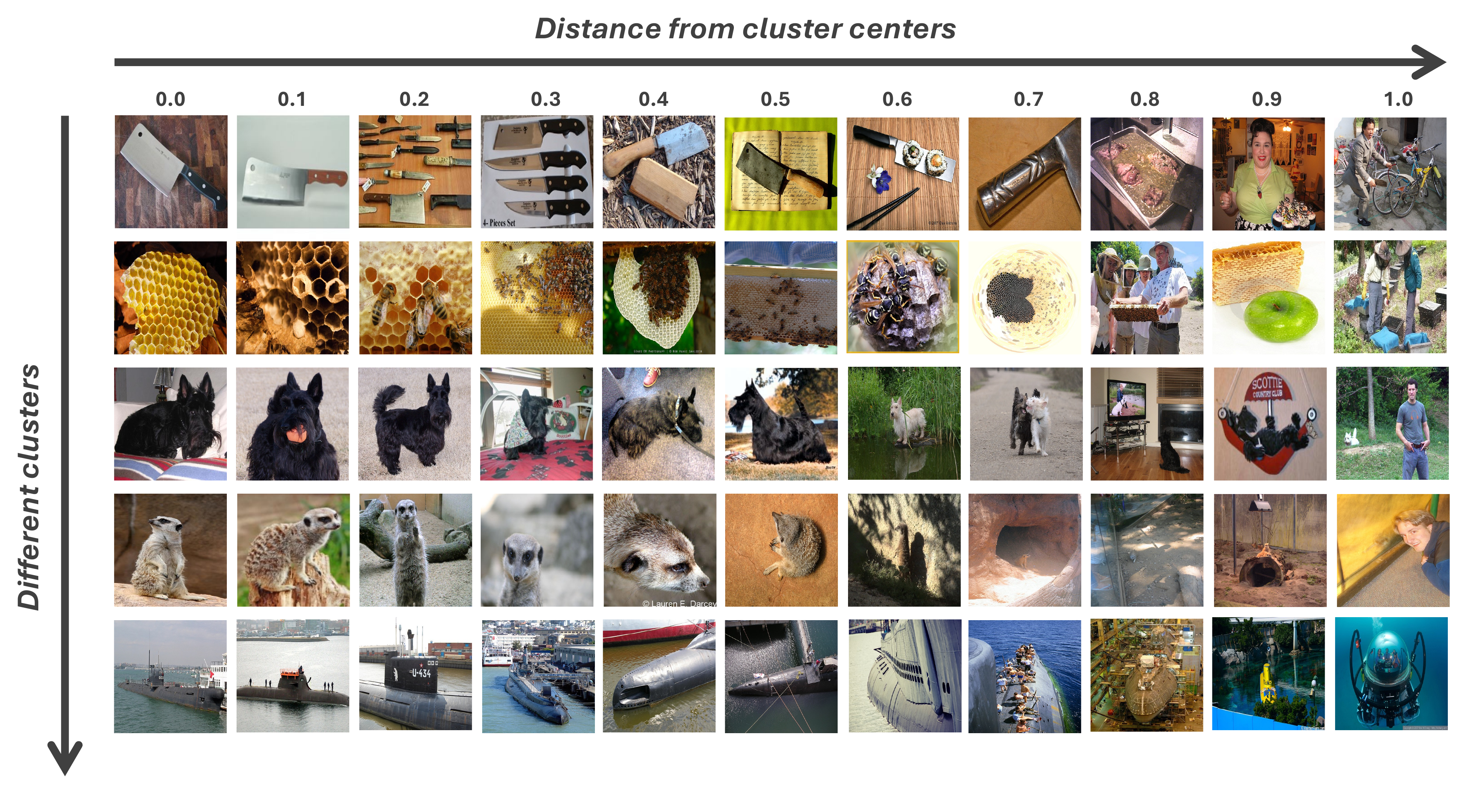}
    \vspace{-1.5em}
    \caption{Visualization of cluster samples arranged by their distances from their respective cluster centers, ranging from nearest to farthest. Each row represents images from the same cluster, while each column indicates the distance of images from their cluster’s centroid.}%
    \vspace{-2ex}
    \label{fig:viz_cluster_samples}
\end{figure*}

\paragraph{Pre-training}
All experiments are conducted on the ImageNet-1K~\cite{deng2009imagenet} dataset, using a ViT-Base as the backbone architecture. 
Following the official MAE~\cite{he2022masked} implementation, images are randomly cropped and resized to 224$\times{}$224 pixels, with random horizontal flipping for data augmentation.
The mask ratio is set to 75\%, and the temperature $\tau$ is annealed from 0.07 to 0.6 over the course of training. Models are optimized using the AdamW optimizer with a batch size of 4096, base learning rate of 1.5e-4, weight decay of 0.05. Linear learning rate warm-up is applied over the first 40 epochs, followed by a cosine learning rate decay schedule.
\vspace{-2ex}

\paragraph{Downstream Evaluation} 
We assess the quality of the learned representations through three complementary evaluation protocols. First, following standard practice in self-supervised learning~\cite{he2021masked,he2020momentum,caron2021emerging,chen2021empirical,lin2024accelerating,wei2025towards}, we evaluate linear probing (LP) performance by training a supervised linear classifier on frozen features extracted from the \texttt{[CLS]} token. This provides a direct measure of the linear separability of the learned representations. Second, we perform full fine-tuning (FT) of the pre-trained model, allowing all parameters to be optimized end-to-end for the downstream classification task. Finally, we evaluate nearest neighbor (NN) classification accuracy using cosine similarity in the feature space, which provides a non-parametric assessment of the learned representations' ability to capture semantic similarities between samples without any additional training.

\begin{table}[t!]
    \centering
    \scalebox{0.95}{\MainExp}
\caption{Comparison of MAE training strategies. Curriculum based on ``Hard-samples'' is adapted from~\cite{sorscher2022beyond}. MAE baseline uses uniform sampling of the training set. Results show NN, LP, and FT accuracy (\%) on ImageNet-1K. }%
    \label{tab:MainExp}
    \vspace{-4ex}
\end{table}

\subsection{Main Results}
We first evaluate our prototype-driven curriculum strategy against alternative approaches for MIM training (\cref{tab:MainExp}). We compare our method against two baselines. The first is a hard-sample curriculum adapted from~\cite{sorscher2022beyond}, which prioritizes non-prototypical images distant from cluster centroids. As shown in~\cite{sorscher2022beyond}, focusing on non-prototypical examples is beneficial for dataset distillation, enabling supervised training on smaller datasets while minimizing performance degradation. We also compare against the standard MAE training regime without any curriculum. All experiments use a ViT-B/16 architecture trained on ImageNet-1k.

The results demonstrate the clear advantages of a prototype-driven curriculum. Most notably, our approach achieves 47.40\% nearest neighbor classification accuracy with 800 epochs of training, substantially outperforming the 800-epoch (30.25\%) MAE baseline. 

Interestingly, while~\cite{sorscher2022beyond} found that non-prototypical (hard) samples need to be preserved when reducing dataset size in supervised learning, our results reveal this strategy to be suboptimal for self-supervised MIM training. The significant performance gap between hard-sample curriculum (62.09\% LP, 82.95\% FT) and our prototypical approach (68.84\% LP, 83.31\% FT) suggests that early exposure to prototypical examples plays a crucial role in developing robust visual representations. This aligns with our hypothesis that building from prototypical to complex examples creates a more stable optimization trajectory in the challenging MIM setting.

\begin{table}[t!]
    \centering
    \scalebox{0.95}{\VarEpoch}
\caption{{\bf Training efficiency comparison.} Performance evaluation of prototype-driven curriculum against standard MAE across different training durations (200 to 800 epochs). The proposed method uses temperature annealing for curriculum learning, while the baseline follows standard uniform sampling. Results show
NN, LP, and FT accuracy (\%) on ImageNet-1K.}%
    \label{tab:VarEpoch}
    \vspace{-4ex}
\end{table}
\vspace{-2ex}
\paragraph{Training Efficiency}
To further investigate the efficiency of our approach, we conducted analyzed model performance across different training durations (\cref{tab:VarEpoch}). We compare our prototypical curriculum against the standard MAE training regime across epochs ranging from 200 to 800, using a ViT-B/16 architecture.

The results reveal that our prototypical curriculum consistently accelerates the learning process. At just 200 epochs, our method achieves 34.92\% nearest neighbor accuracy and 63.74\% linear probing performance, already surpassing the 800-epoch baseline (30.25\% NN, 64.25\% LP). This early advantage becomes more pronounced with additional training: at 400 epochs, our approach (38.09\% NN, 66.34\% LP) already exceeds the performance of the 800-epoch baseline (30.25\% NN, 67.77\% LP).

Overall, prototype-driven curriculum learning enables more efficient and effective representation learning in the MIM framework.
\vspace{-2ex}

\begin{figure}[t]
    \centering
    \vspace{-3ex}
    \includegraphics[width=1.0\linewidth]{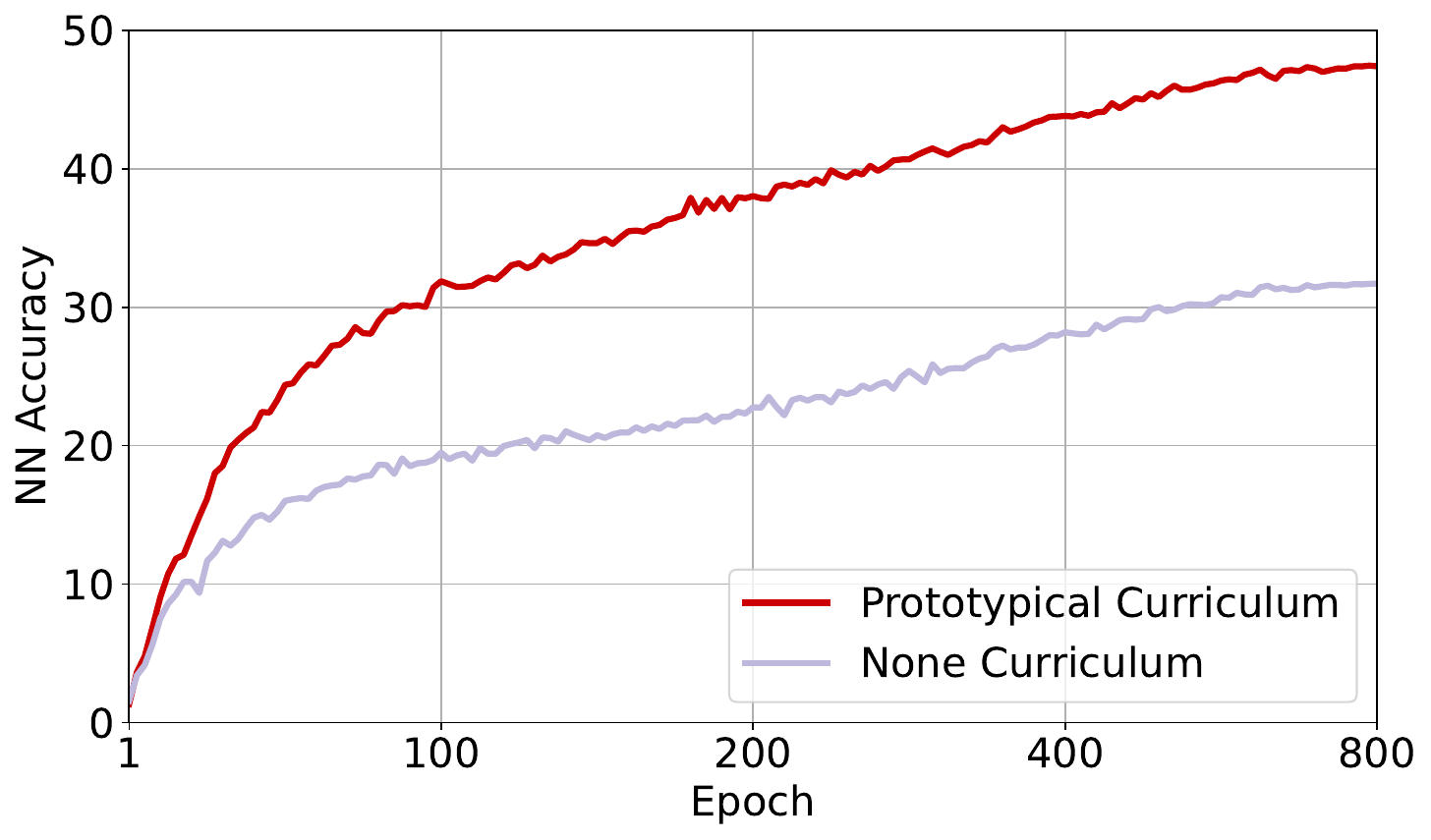}
    \vspace{-5ex}
    \caption{{\bf Training curve.} Nearest-neighbor accuracy throughout training for an MAE model trained with and without a prototype-driven curriculum.}%
    \label{fig:training_curve}
    \vspace{-3ex}
\end{figure}
\paragraph{Training Dynamics}
We also analyzed the training dynamics using nearest-neighbor accuracy as a proxy for representation quality. We train two MAE models with identical ViT-Base architectures on ImageNet-1k for 800 epochs: one with our proposed curriculum and one with the standard training procedure.

Our experiments demonstrate that incorporating a prototype-driven curriculum significantly improves the training dynamics of MAE models. As shown in \cref{fig:training_curve}, the model trained with our curriculum exhibits notably faster learning. The benefits become apparent very early in training, with curriculum training achieving the same level of performance of the baseline model after only 100 epochs (8$\times$ faster), and outperforming it by a large margin as the training progresses.
\vspace{-2ex}

\paragraph{Few-shot Performance}
To further validate the quality and effectiveness of the representations learned by our accelerated pretraining method, we perform linear probing evaluations under a few-shot scenario. Specifically, we assess the performance of our pretrained models on eight fine-grained datasets, using a 5-shot setup where only five labeled examples per class are available for training the linear classifier. This evaluation tests the generalization capabilities and data efficiency of the learned representations. Results are summarized in \cref{tab:FewShotExp}, which clearly demonstrate that our method, even at significantly fewer training epochs (200 and 400 epochs), achieves comparable or superior performance compared to the baseline trained for 1600 epochs, highlighting both efficiency and effectiveness with up to an 8$\times$ speedup.

\begin{table*}[t!]
    \centering
    \resizebox{\textwidth}{!}{
       \begin{tabular}{l|cccccccc|c}
           \hline
           & \bf CUB-200~\cite{wah2011caltech} & \bf SUN397~\cite{xiao2010sun} & \bf OxfordPets~\cite{parkhi12a} & \bf Caltech101~\cite{fei2004learning} & \bf UCF101~\cite{soomro2012ucf101} & \bf DTD~\cite{cimpoi14describing} & \bf Flowers102~\cite{Nilsback08} & \bf Food101~\cite{bossard14} & \bf AVG \\
           \hline
           Baseline (1600 epoch) & 14.42 {\scriptsize ±0.38} & 30.20 {\scriptsize ±0.61} & 39.76 {\scriptsize ±2.43} & 74.81 {\scriptsize ±2.11} & 49.06 {\scriptsize ±0.67} & 44.47 {\scriptsize ±1.89} & 64.48 {\scriptsize ±0.85} & 16.97 {\scriptsize ±0.42} & 41.77 \\
           Ours (200 epoch) & 14.35 {\scriptsize ±0.41} & 30.15 {\scriptsize ±1.03} & 43.79 {\scriptsize ±1.75} & 77.42 {\scriptsize ±1.51} & 48.15 {\scriptsize ±1.05} & 42.81 {\scriptsize ±1.79} & 64.13 {\scriptsize ±0.71} & 16.64 {\scriptsize ±0.05} & 42.18 \\
           Ours (400 epoch) & \textbf{18.46 {\scriptsize ±0.40}} & \textbf{34.33 {\scriptsize ±1.24}} & \textbf{56.97 {\scriptsize ±2.20}} & \textbf{82.02 {\scriptsize ±1.86}} & \textbf{52.35 {\scriptsize ±1.47}} & \textbf{45.59 {\scriptsize ±1.20}} & \textbf{68.99 {\scriptsize ±0.53}} & \textbf{19.11 {\scriptsize ±0.36}} & \textbf{47.23} \\
           \hline
       \end{tabular}}
       \vspace{-2ex}
       \caption{5-shot linear probing accuracy (\%) on eight fine-grained datasets. Mean and standard deviation reported over three runs. Our accelerated method (200 and 400 epochs) achieves comparable or improved performance compared to the baseline method (1600 epochs), highlighting a training speedup of up to 8×.}
    \vspace{-6pt}
    \label{tab:FewShotExp}
\end{table*}

\begin{table*}[t!]
    \begin{minipage}{0.35\linewidth}
        \centering
        \vspace{4em}
        \TemperatureExp{}
        \caption{{\bf Impact of temperature on training.} Comparison between fixed and annealed temperature sampling. Lower temperatures ($\tau$) favor prototypical examples; $\tau=\infty$ represents uniform sampling. $|\gD_\tau| / |\gD|$ denotes effective dataset size. Annealed sampling adjusts $\tau$ from 0.07 to 0.6.}
        \label{tab:TemperatureExp}
    \end{minipage}
    \hfill
    \begin{minipage}{0.6\linewidth}
        \begin{minipage}{0.48\linewidth}
            \centering
            \scalebox{0.8}\AlgorithmExp{}
            \subcaption{Feature space.}%
            \label{tab:AlgorithmExp}
            \vspace{1em}
            \scalebox{0.85}{\ArchitectureExp{}}
            \subcaption{DiNO Backbone.}%
            \label{tab:ArchitectureExp}
        \end{minipage}
        \hfill
        \begin{minipage}{0.48\linewidth}
            \centering
            \scalebox{0.85}{\NumClusterExp{}}
            \subcaption{Number of clusters. 978 is computed via DB~\ref{para:DB}.}%
            \label{tab:NumClusterExp}
        \end{minipage}
        \caption{Empirical study on prototype identification. We evaluate the impact of (a) feature space, (b) model architectures, and (c) number of clusters. We further compare self-supervised DiNO clusters against ``Oracle'' clusters which use ground truth labels to define the centroids. Unless otherwise specified, experiments use DiNO features from a ViT-Base/16 backbone with 1000 clusters. All models are trained for 800 epochs on ImageNet-1K.}%
        \label{tab:combined}
    \end{minipage}
    \vspace{-3ex}
\end{table*}

\subsection{Impact of Temperature Annealing}
Temperature annealing enables the model to transition from prototypical to diverse examples during training.
To examine its role in our curriculum strategy, we compared it against fixed temperature $\tau$ schedules, with temperatures ranging from 0.01 to infinity in~\cref{tab:TemperatureExp}. Constant temperature effectively creates static training distributions, with lower temperatures prioritizing more prototypical examples.

The results demonstrate that static temperature choices face an inherent trade-off. At very low temperatures ($\tau$ = 0.01), the model sees only the most prototypical examples but achieves poor performance (5.69\% NN, 18.22\% LP), likely due to limited exposure to variations in the data. As temperature increases, performance improves up to an optimal point around $\tau$ = 0.2 (41.57\% NN, 67.71\% LP), before gradually declining at higher temperatures. Interestingly, the optimal static temperature of $\tau=0.2$ outperforms the baseline MAE model which is trained under uniform sampling (\ie, $\tau=\infty$). This demonstrates that even a static bias towards prototypical examples outperforms standard uniform sampling, suggesting that prototypical examples play a crucial role in MIM training.

However, the optimal static temperature still underperforms temperature annealing (47.40\% NN, 68.80\% LP). These findings show that a curriculum transitioning from prototypical to diverse examples is beneficial for effective MIM training. Rather than settling for a fixed compromise between prototypicality and diversity, temperature annealing allows the model to first establish robust representations from clear prototypes before gradually incorporating more challenging variations.


\subsection{Impact of Prototype Identification}
To study the impact of prototype identification on our curriculum learning approach, we conducted a comprehensive study examining three key aspects: the choice
of feature space, backbone architecture, and clustering granularity. All experiments were conducted on ImageNet-1k using a ViT-B/16 backbone architecture.

First, we investigated the impact of different representation spaces (\cref{tab:AlgorithmExp}). Prototypes selected from DiNO features yielded superior performance (NN: 40.15\%, LP: 68.73\%) compared to prototypes obtained from both an untrained model (NN: 32.63\%, LP: 65.65\%) and an MAE-pretrained model (NN: 34.14\%, LP: 66.60\%). Interestingly, DINO prototypes even slightly outperformed supervised pre-training (NN: 39.70\%, LP: 68.09\%), showing that supervised pre-training is not required for effective prototype identification. 
In fact, even curricula derived from clustering in a low-level feature space, such as SIFT, already performed well  (NN: 36.85\%, LP: 69.17\%). Large pre-trained models, which might be expensive to train, are not required to establish the curriculum.

\cref{tab:ArchitectureExp} shows that the size of the backbone architecture is not a critical factor in prototype identification. In fact, slightly better
downstream performance is achieved with the smaller ViT-Small model (NN: 41.05\%, LP: 69.23\%) compared to the larger ViT-Base/16 model (NN: 40.15\%, LP:
68.73\%). 

Finally, our study of clustering granularity (\cref{tab:NumClusterExp}) reveals an optimal range for the number of centroids. Performance peaks at 1000 clusters
(NN: 41.05\%, LP: 69.23\%), with degradation observed at both lower (500 clusters) and higher (2000 clusters) granularities. Notably, the data-driven approach
to determining cluster count through the Davies-Bouldin index (978 clusters) achieves comparable performance to the manually tuned optimum. Interestingly,
comparing our self-supervised clustering approach to an oracle baseline using ground truth labels for centroid definition shows that our method actually
outperforms the oracle in nearest neighbor classification (41.05\% vs 37.83\%). This surprising result suggests that our prototype identification method
captures underlying visual patterns that may be more fundamental than human-defined categorical boundaries, potentially offering a more nuanced representation
of visual similarities.

These ablation studies collectively demonstrate the robustness of our approach and provide practical insights for implementation. The results support our
hypothesis that prototype-driven curriculum learning benefits from (1) self-supervised feature spaces that capture rich semantic relationships, (2)
architectures that preserve fine-grained spatial information, and (3) careful optimization of prototype granularity.%
\vspace{-4pt}

\section{Conclusion}%
\label{sec:conclusion}
In this work, we introduced a prototype-driven curriculum learning approach for Masked Image Modeling that structures the learning process to progress from prototypical to more complex examples. As the computational demands of training large vision models continue to grow, improving training efficiency becomes increasingly critical for sustainable AI development. Through extensive experiments, we demonstrated that our method substantially outperforms standard MAE training and alternative curriculum strategies with significantly reduced training time. These improvements stem from our carefully designed temperature annealing strategy, which enables the model to first establish robust representations from clear prototypes before gradually incorporating more challenging variations.

While we focused our evaluation on ImageNet-1K, our findings have promising implications for scaling MAE methods to larger and more complex datasets. The demonstrated efficiency gains suggest that prototype-driven curricula could be particularly valuable in scenarios where computational resources are constrained or dataset complexity poses training challenges. Future work could explore the application of our approach to massive-scale datasets, where the benefits of structured learning from prototypical examples may be even more pronounced. Additionally, investigating how prototype selection strategies could be adapted for diverse visual domains could further broaden the impact of this approach.

{
    \small
    \bibliographystyle{style-cvpr/ieeenat_fullname}
    \bibliography{refs}
}



\clearpage
\appendix 














\section*{Supplementary Material}


The supplementary material provides a detailed configuration used for downstream tasks, as well as additional visualizations of the prototypical scores obtained across different feature spaces.

\paragraph{Training Configuration for Downstream Tasks.}
End-to-end fine-tuning and linear probe settings are summarized in
Tables~\ref{tab:impl_mae_finetune} and~\ref{tab:impl_mae_linear}, respectively.
We adopt the linear learning rate scaling rule: \(\text{lr} = \text{base lr} \times
\text{batchsize} / 256\), consistent with the original implementation.

\begin{figure*}[t]
    \centering
    \setcounter{subfigure}{0}
    \begin{subfigure}[t]{0.92\textwidth}
        \centering
        \includegraphics[width = 0.92\textwidth]{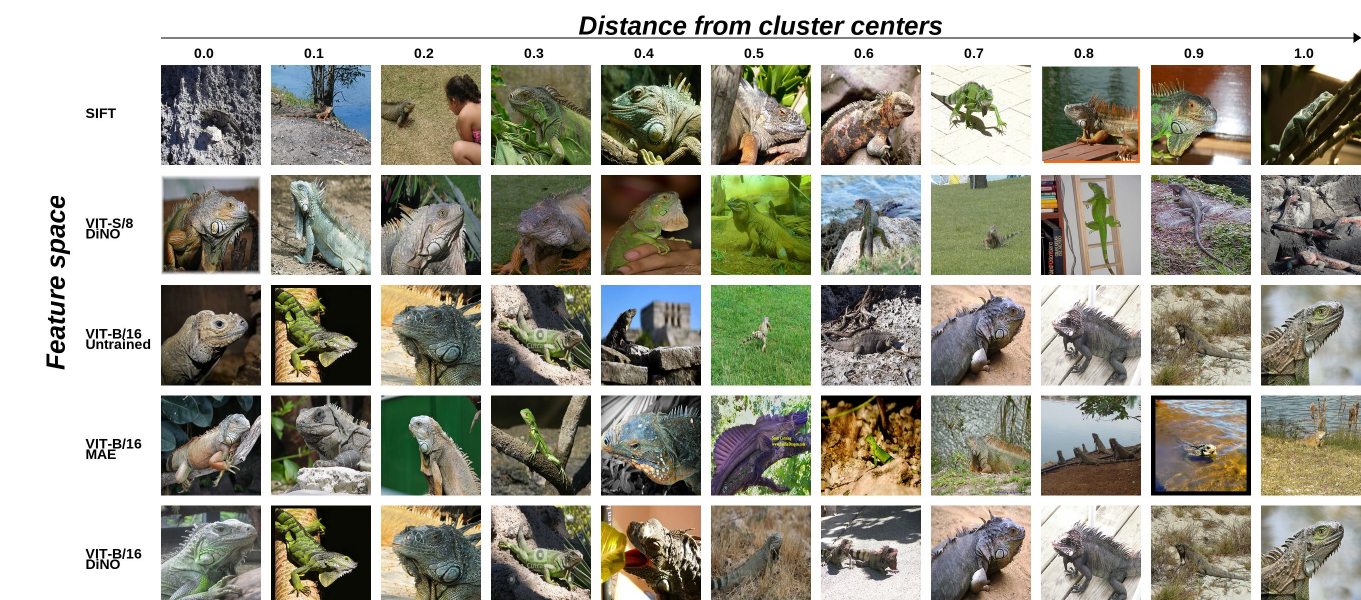}
        \caption{Visualization for synsets n01677366 (iguana).}
        \label{fig:viz_feature_space_class_a}
    \end{subfigure}
    \vspace{1em}
    \begin{subfigure}[t]{0.92\textwidth}
        \centering
        \includegraphics[width = 0.92\textwidth]{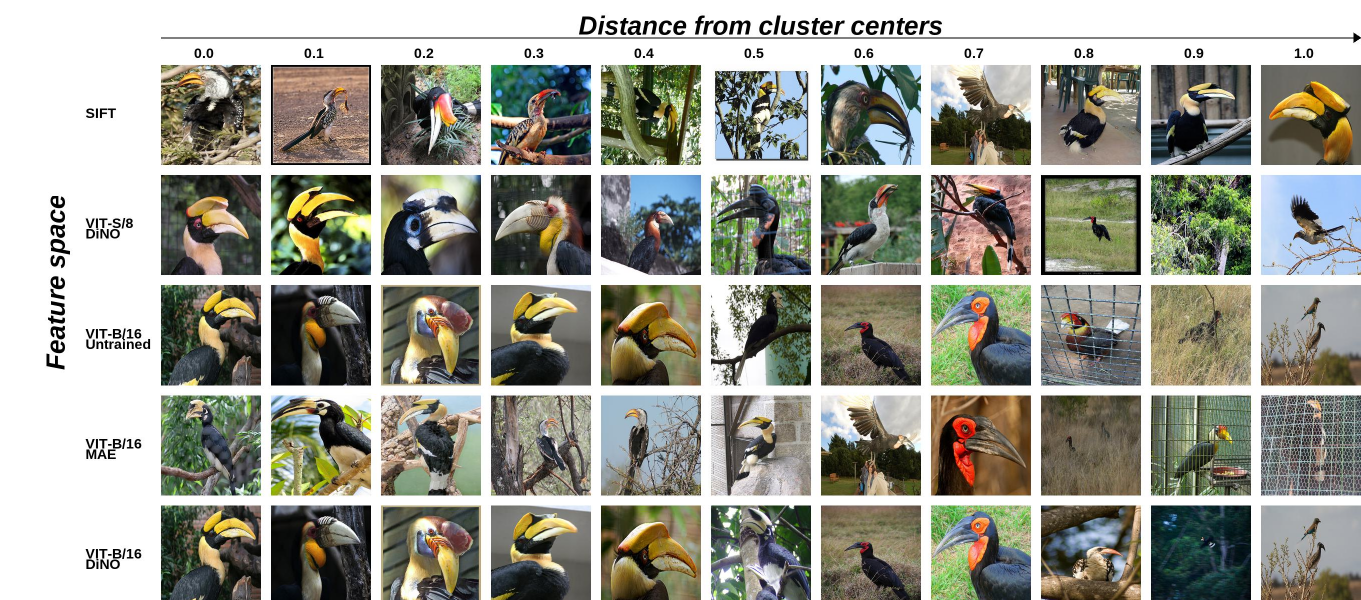}
        \caption{Visualization for synsets n01829413 (hornbill).}
        \label{fig:viz_feature_space_class_b}
    \end{subfigure}
    \vspace{1em}
    \begin{subfigure}[t]{0.92\textwidth}
        \centering
        \includegraphics[width = 0.92\textwidth]{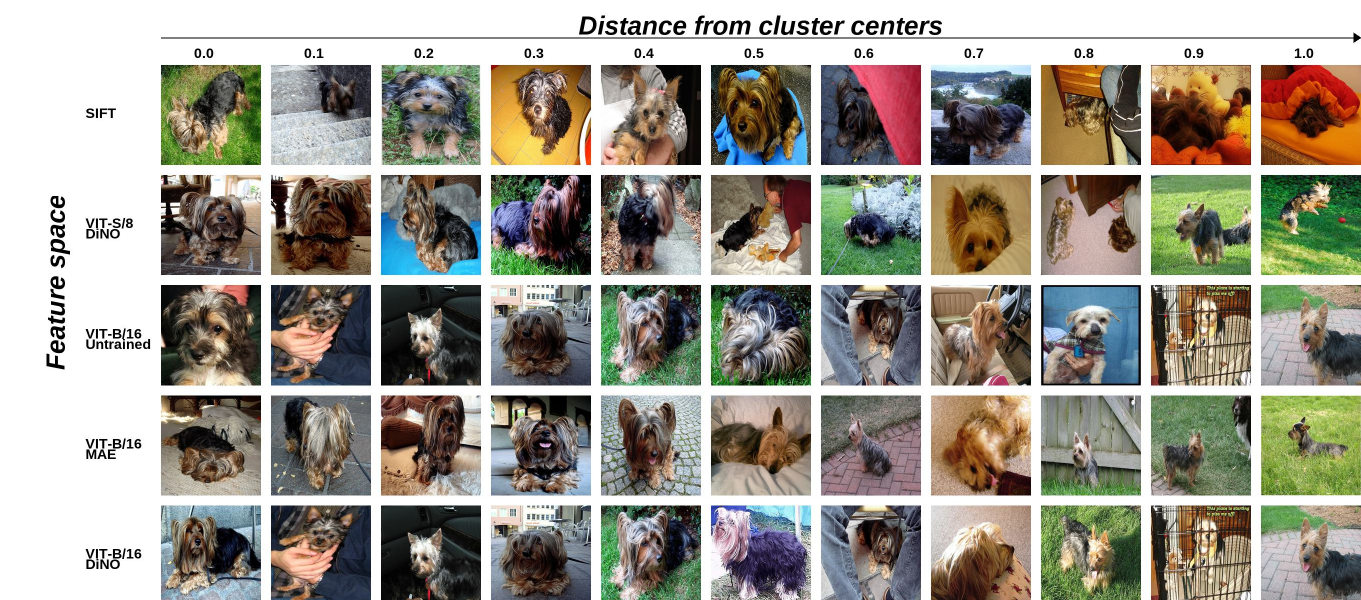}
        \caption{Visualization for synsets n02097658 (silky terrier).}
        \label{fig:viz_feature_space_class_c}
    \end{subfigure}
    \caption{Visualization of cluster samples arranged by their distances from their respective cluster centers (Part 1 of 3). Each row represents images from the same model, while each column indicates the distance of images from their cluster's centroid.}
    \label{fig:viz_feature_space_combined_1}
\end{figure*}

\paragraph{Visualization of various Feature Spaces.}
We visualize the images with various distances from their respective cluster centroid in different feature spaces. The visualization is shown in Figure~\ref{fig:viz_feature_space_combined_1}.
The DINO representations space (regardless of the size of the backbone architecture) consistently places prototypical views of each object closer to the cluster centroids. Results using low-level features, such as MAE or SIFT, appear to be less organized.

\begin{table}[h]
\begin{tabular}{r|l}
\bf Parameter & \bf Value \\
\shline
Optimizer & AdamW \\
Training epochs & 100 \\
Batch size & 1024 \\
Base learning rate (LR) & 1e-3 \\
LR schedule & cosine decay \\
LR Warm-up epochs & 5 \\
Weight decay & 0.05 \\
Optimizer momentum & $\beta_1, \beta_2{=}0.9, 0.999$ \\
Layer-wise LR decay  & 0.75 \\
Augmentation & RandAug (9, 0.5) \\
Label smoothing  & 0.1 \\
Mixup  & 0.8 \\
CutMix  & 1.0 \\
Drop path  & 0.1 \\
\end{tabular}
\caption{\textbf{End-to-end fine-tuning setting.}}
\label{tab:impl_mae_finetune}
\end{table}

\begin{table}[h]
\begin{tabular}{r|l}
\bf Parameter & \bf Value \\
\shline
Optimizer & LARS  \\
Training epochs & 90 \\
Batch size & 16384 \\
Base learning rate (LR) & 0.1 \\
LR schedule & cosine decay \\
LR Warm-up epochs & 10 \\
Weight decay & 0 \\
Optimizer momentum & 0.9 \\
Augmentation & RandomResizedCrop \\
\end{tabular}
\caption{\textbf{Linear probing setting.}
\label{tab:impl_mae_linear}}
\end{table}

\begin{figure*}[t]
    \centering
    \setcounter{subfigure}{3}
    \begin{subfigure}[t]{0.92\textwidth}
        \centering
        \includegraphics[width = 0.92\textwidth]{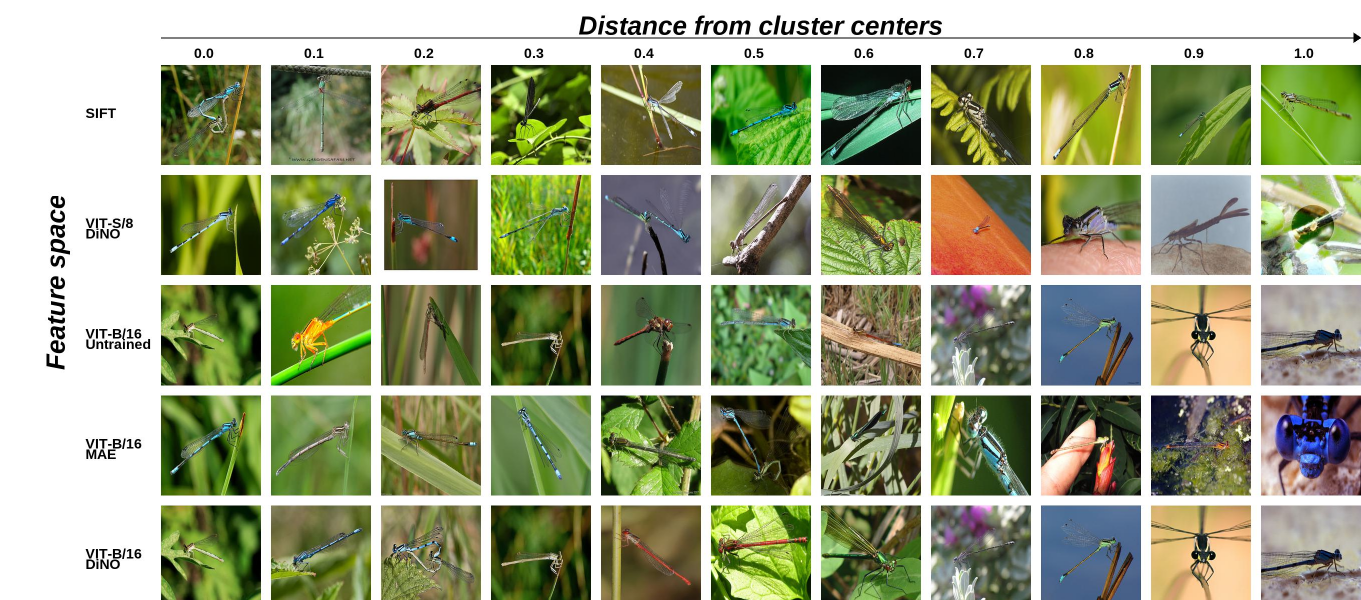}
        \caption{Visualization for synsets n02268853 (damselfly).}
        \label{fig:viz_feature_space_class_d}
    \end{subfigure}
    \vspace{1em}
    \begin{subfigure}[t]{0.92\textwidth}
        \centering
        \includegraphics[width = 0.92\textwidth]{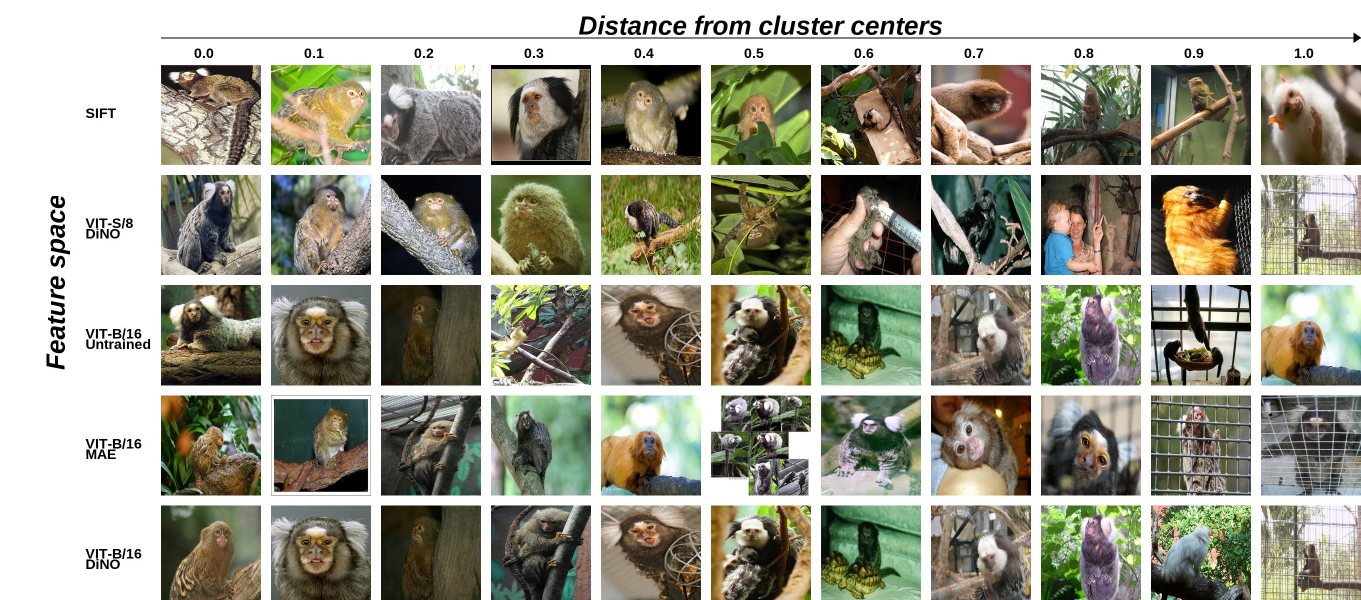}
        \caption{Visualization for synsets n02490219 (marmoset).}
        \label{fig:viz_feature_space_class_e}
    \end{subfigure}
    \vspace{1em}
    \begin{subfigure}[t]{0.92\textwidth}
        \centering
        \includegraphics[width = 0.92\textwidth]{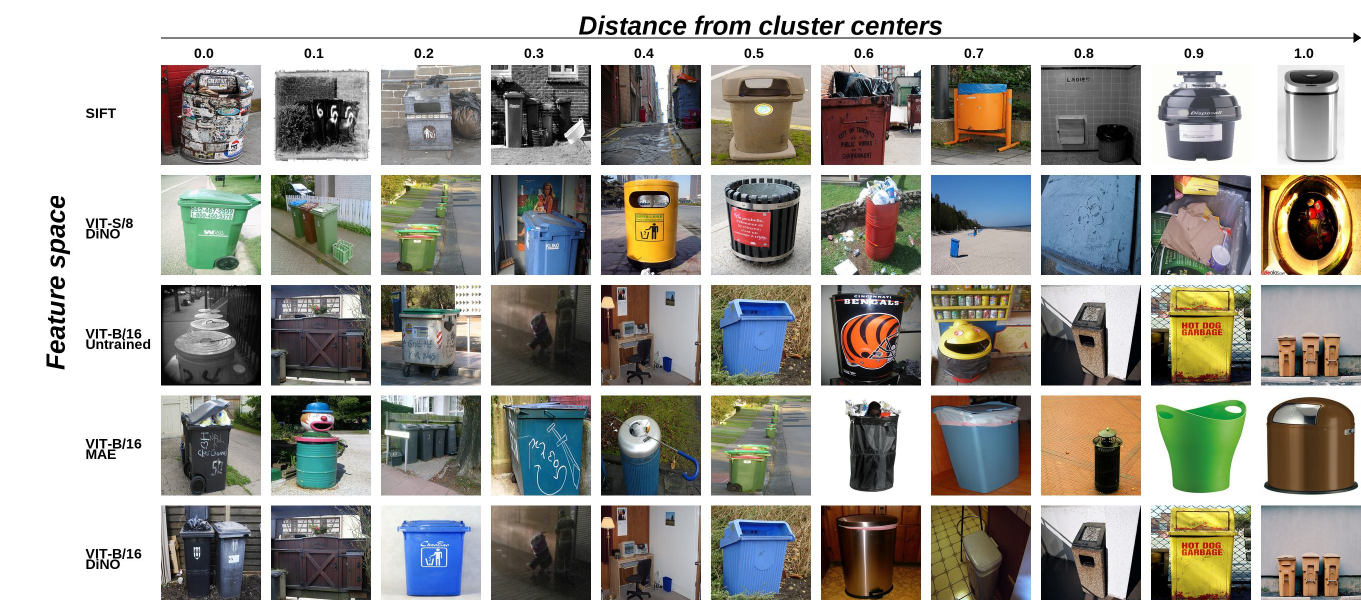}
        \caption{Visualization for synsets n02747177 (trash can).}
        \label{fig:viz_feature_space_class_f}
    \end{subfigure}
    \caption{Visualization of cluster samples arranged by their distances from their respective cluster centers (Part 2 of 3). Each row represents images from the same model, while each column indicates the distance of images from their cluster's centroid.}
    \label{fig:viz_feature_space_combined_2}
\end{figure*}

\begin{figure*}[t]
    \centering
    \setcounter{subfigure}{6}
    \begin{subfigure}[t]{0.92\textwidth}
        \centering
        \includegraphics[width = 0.92\textwidth]{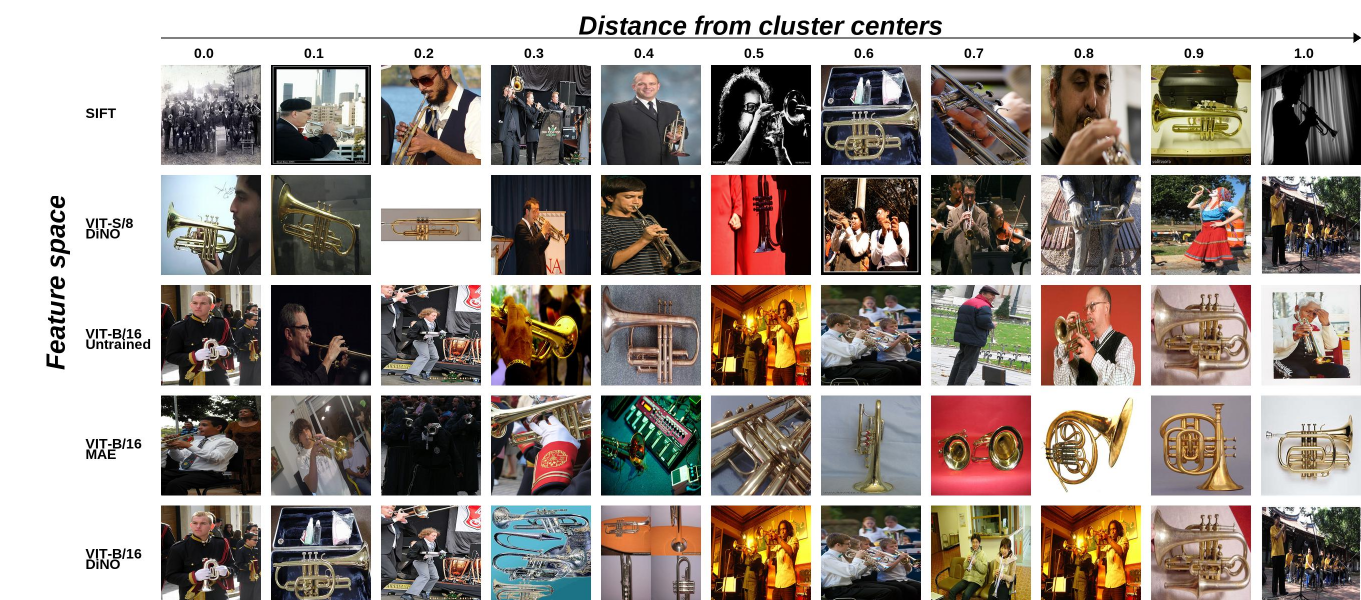}
        \caption{Visualization for synsets n03110669 (cornet).}
        \label{fig:viz_feature_space_class_g}
    \end{subfigure}
    \vspace{1em}
    \begin{subfigure}[t]{0.92\textwidth}
        \centering
        \includegraphics[width = 0.92\textwidth]{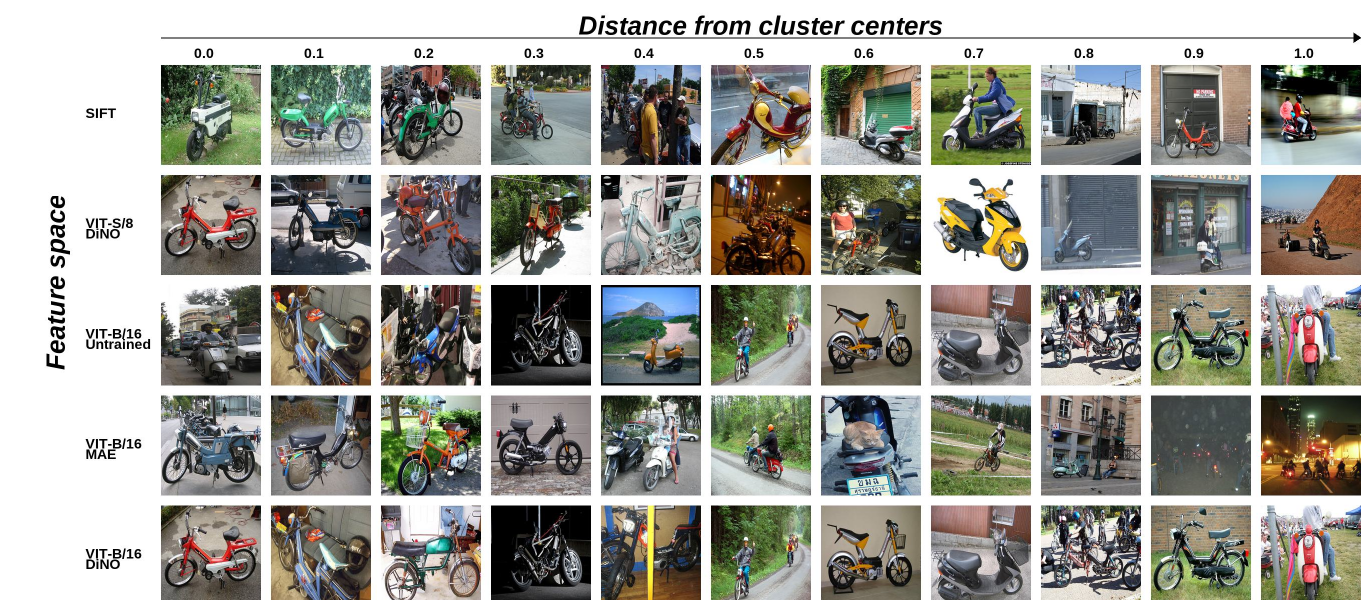}
        \caption{Visualization for synsets n03785016 (moped).}
        \label{fig:viz_feature_space_class_h}
    \end{subfigure}
    \vspace{1em}
    \begin{subfigure}[t]{0.92\textwidth}
        \centering
        \includegraphics[width = 0.92\textwidth]{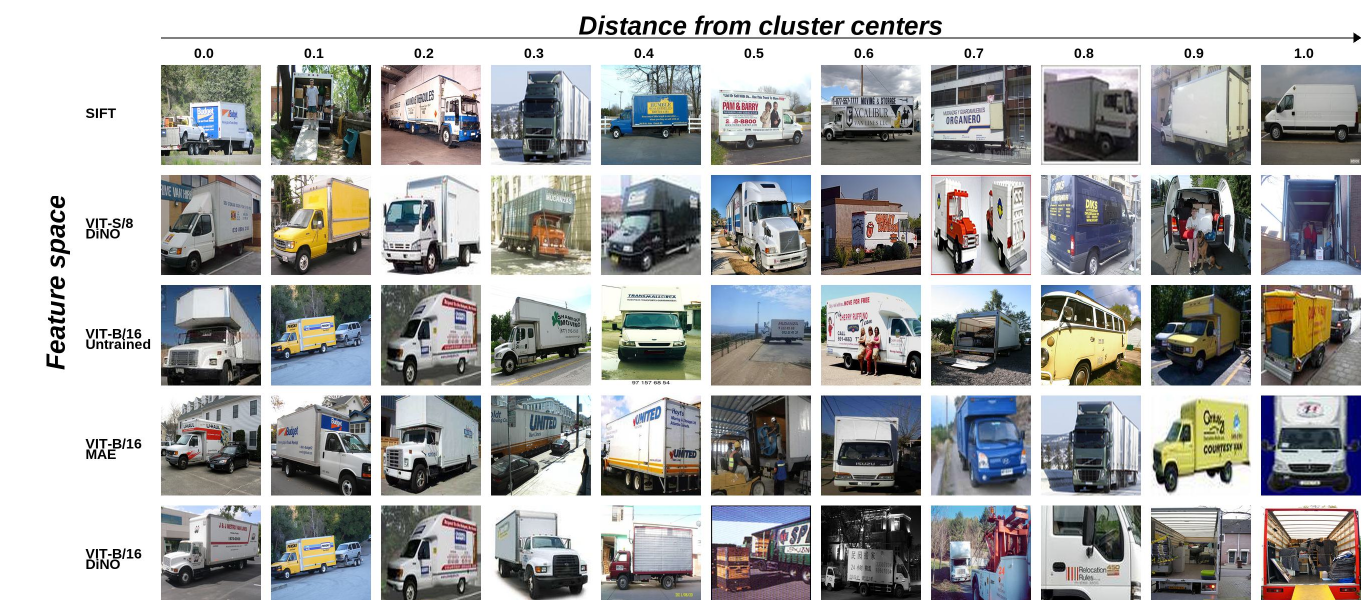}
        \caption{Visualization for synsets n03796401 (moving van).}
        \label{fig:viz_feature_space_class_i}
    \end{subfigure}
    \caption{Visualization of cluster samples arranged by their distances from their respective cluster centers (Part 3 of 3). Each row represents images from the same model, while each column indicates the distance of images from their cluster's centroid.}
    \label{fig:viz_feature_space_combined_3}
\end{figure*}


\end{document}